\documentclass{article}
\usepackage{spconf,amsmath,graphicx}
\usepackage{amsfonts,amssymb}
\usepackage{algorithm}
\usepackage[noend]{algpseudocode}
\usepackage{caption}
\usepackage{listings}
\usepackage[x11names]{xcolor}
\usepackage{color}
\usepackage{xspace}
\usepackage{bbm}
\usepackage{dsfont}

\makeatletter
\def\BState{\State\hskip-\ALG@thistlm}
\makeatother

\renewcommand{\Re}[1]{\mbox{$\mathbbm{R}^{#1}$}}

\def\indi#1{\mbox{$\mathds{1}_{#1}$}}

\newdimen\origiwspc
\newdimen\origiwstr
\origiwspc=\fontdimen2\font %
\origiwstr=\fontdimen3\font %

\def\eg{\emph{e.g.}}
\def\tcell{\mbox{T--cell}}
\def\tcells{\mbox{T--cells}}

\title{Multiclass weighted loss for instance segmentation\\ of cluttered cells}
\name{{\parbox[c]{0.9\textwidth}{\centering Fidel A. Guerrero-Pe\~{n}a$^{1*}$, Pedro D. Marrero Fernandez$^1$, Tsang Ing Ren$^1$,\\
Mary Yui$^2$, Ellen Rothenberg$^2$, Alexandre Cunha$^{3*}$
  \thanks{\footnotesize\fontdimen2\font=0.35ex
We thank financial support from the Brazilian funding agencies
  FACEPE, CAPES and CNPq (FAG, PF, TIR), from the Beckman Institute at Caltech
    to the Center for Advanced Methods in Biological Image Analysis (AC), and
    thank the IBM Matching Grants Program for computer donation (AC).
    $^*$Corresponding authors: \texttt{fagp@cin.ufpe.br,cunha@caltech.edu.}}}}}

\address{
$^1$Centro de Inform\'atica, Universidade Federal de Pernambuco, Brazil\\
$^2$Division of Biology and Biological Engineering, California Institute of Technology, USA\\
$^3$Center for Advanced Methods in Biological Image Analysis, California Institute of Technology, USA
}

\begin{document}
\ninept
\maketitle
\begin{abstract}
  We propose a new multiclass weighted loss function for instance segmentation
  of cluttered cells.  We are primarily motivated by the need of developmental
  biologists to quantify and model the behavior of blood \tcells\ which might
  help us in understanding their regulation mechanisms and ultimately help
  researchers in their quest for developing an effective immunotherapy cancer
  treatment. Segmenting individual touching cells in cluttered regions is
  challenging as the feature distribution on shared borders and cell foreground
  are similar thus difficulting discriminating pixels into proper classes.  We
  present two novel weight maps applied to the weighted cross entropy loss
  function which take into account both class imbalance and cell geometry.
  Binary ground truth training data is augmented so the learning model can
  handle not only foreground and background but also a third touching class.
  This framework allows training using \mbox{U-Net}. Experiments with our
  formulations have shown superior results when compared to other similar
  schemes, outperforming binary class models with significant improvement of
  boundary adequacy and instance detection. We validate our results on manually
  annotated microscope images of \tcells.
\end{abstract}

\section{Introduction}
\label{sec:intro}

\begin{figure}[t]
  \begin{center}
  \includegraphics[width=\columnwidth]{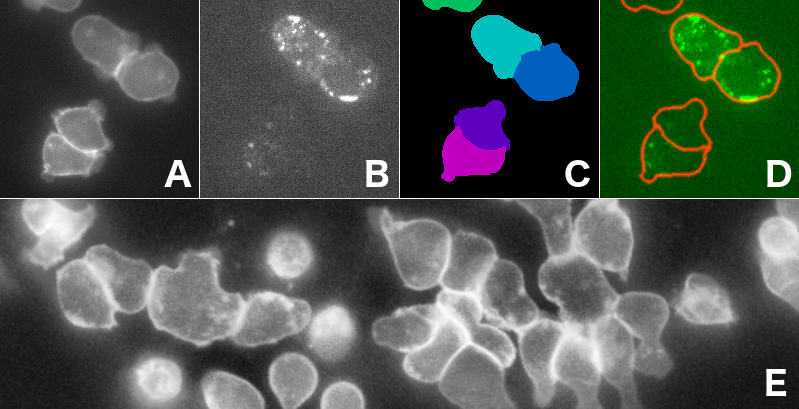}
  \caption{\footnotesize
  We show in (A) cells marked by the
    {\it mTomato} fluorophore. Their corresponding signal of interest, {\it
    CD25}, which changes over time, is expressed in some cells (B). Our
    goal is to segment individual cells, as shown in (C), and colocalize {\it
    CD25} to measure its concentration within each cell (D) and consequently
    count how many cells are active at any given time. In this illustration, the top
    two cells are fully active as reported by their high {\it CD25} content. Colored masks in (C) are for illustration purpose only.
    A typical cluttering of {T--cells} is presented on panel E, which shows the maximum
    intensity projection of a few slices of a widefield stack.
    }
  \vspace{-2.0mm}
    \color{lightgray}\rule{1.0\columnwidth}{0.2mm}
  \label{fig:cells}
  \end{center}\vspace{-1.3cm}
\end{figure}

It is not fully understood how blood stem cells differentiate over time to
generate all blood cell types in the body nor what are the mechanisms that
drive their specialization. \tcells\ are descendants of blood stem cells
with an important role in emerging immunotherapy cancer treatments
\cite{rosenberg2015}.  We are particularly
interested in determining how decisions are made by individual progenitor
\tcells\ under controlled environmental conditions \cite{rothenberg2008}.
To carry out experiments, individual \tcells\ are isolated in microwells
where they grow and proliferate for five or six days.  Multiple cell divisions
occur in each microwell leading to a dense cell population originated from a
single cell. Multichannel images are acquired at intervals to follow cell
development which can then be quantified by analyzing fluorescent signals
expressing specific markers of differentiation.  Segmenting individual cells is
necessary to measure signal activation per cell and to count how many cells are
active over time (see Fig.\ref{fig:cells}).

\begin{figure*}[!t]
  \begin{center}
  \centering
\setlength{\tabcolsep}{1pt}
\begin{tabular}{ccc}
\includegraphics[width=.14\linewidth]{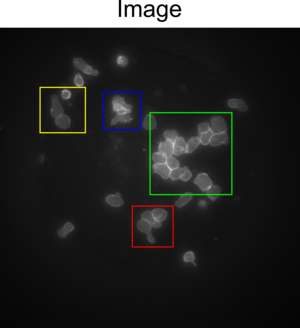}&
\includegraphics[width=.41\linewidth]{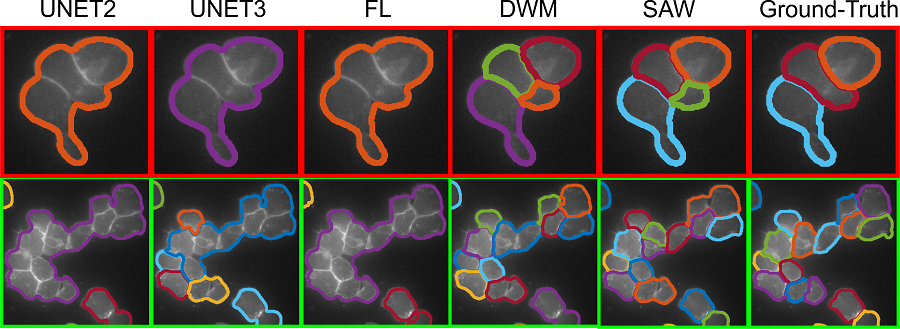}&
\includegraphics[width=.41\linewidth]{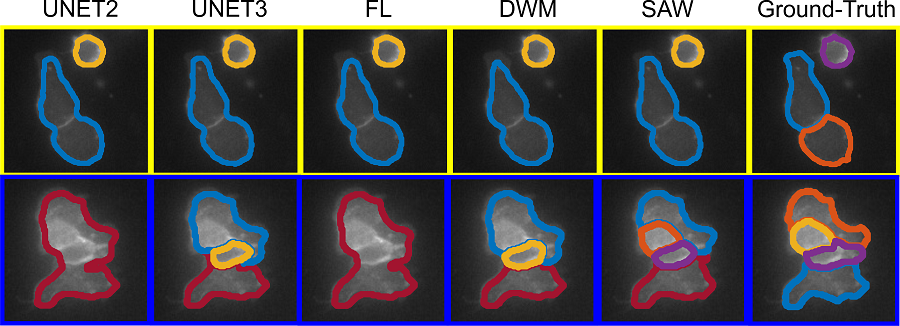}\\

\includegraphics[width=.14\linewidth]{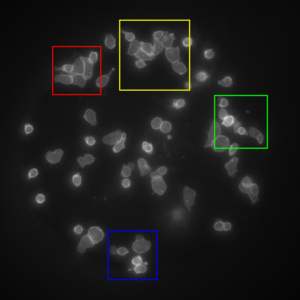}&
\includegraphics[width=.41\linewidth]{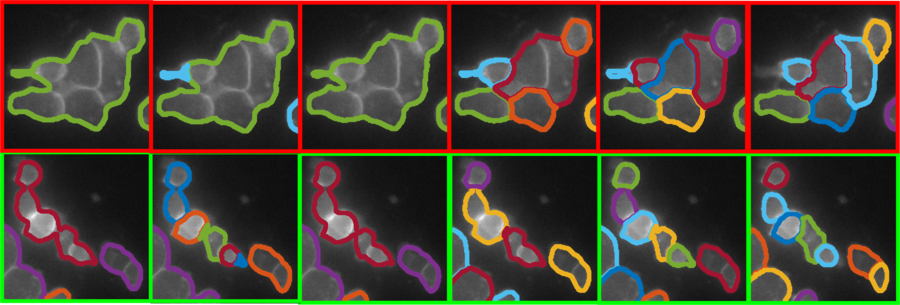}&
\includegraphics[width=.41\linewidth]{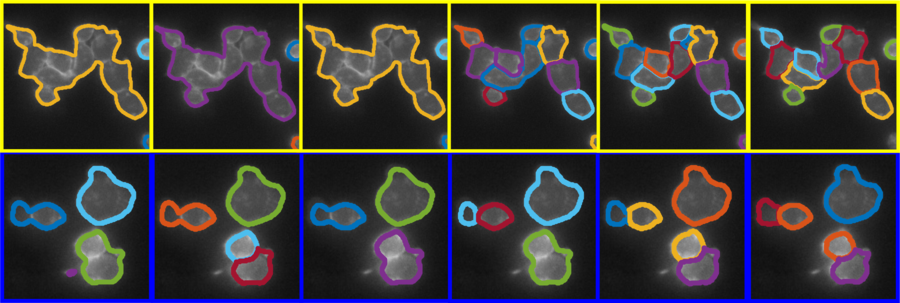}\\
\end{tabular}
  \caption{\footnotesize Results of instance segmentation obtained with UNET2,
  UNET3, FL, DWM, and SAW and ground truth delineations for eight regions of
  two images. Results are for the best combination of $\gamma_1$ and
  $\gamma_2$ thresholds. Note that the two cells on top right set cannot be resolved by any
  method. This might be due to weak signal on a very small edge or lack of
  sufficient training data. Our methods render close to optimal solutions for
  almost all training and testing data. We expect enhancements, for all
  methods, when larger training data is available.  Contour colors are only to
  illustrate separation of cells.}
  \vspace{-2.0mm}
  \noindent{\color{lightgray}\rule{0.98\linewidth}{0.2mm}} \label{fig:res1}
  \end{center}\vspace{-0.8cm}
\end{figure*}

The difficulties are in segmenting adjoining cells. These can take any shape,
when cluttered or isolated, and their touching borders have nonuniform
brightness and patterns defeating classical segmentation approaches. Weak
boundaries are also troubling (see Fig.\ref{fig:cells} and also
Fig.\ref{fig:res1}). Furthermore, the total pixel count on adjoining borders is
considerably smaller than the pixel count for the entire image which
contributes to numerical optimization difficulties when training a neural network with
imbalanced data \cite{he2009} and without a properly calibrated loss function. The situation
is exacerbated in large clusters where cells might
overlap making it difficult, even for the trained eye, to locate cell
contours. We approach these difficulties by adopting a loss function with
pixel-wise weights, following \cite{ronneberger2015u}, that take into
account not only the location and length of touching borders but also the
geometry of cell contours.

The problem of segmenting cells with difficult boundaries has been addressed by
others.  Long {\it et al.} \cite{long2015fully} proposed a Fully Convolutional
Network (FCN) which improved the image-level classification in a Convolutional
Neural Network (CNN) to a pixel-level classification. This allowed segmentation
maps to be generated for images of any size and it was much faster compared to
the then prevalent patch classification approach.  In the same year,
Ronneberger {\it et al.}  \cite{ronneberger2015u} introduced U-Net to segment
biomedical images, a FCN encoder-decoder type of architecture, together with a
weighted cross entropy loss function. This network was a breakthrough,
achieving remarkable results in segmenting biomedical images, from cells to
organs. We have opted to use U-Net due to its proven success but we employ
different per pixel weights in the loss function.

Browet {\it et al.} \cite{browet2016}, working with mouse embryo cells,
estimated pixel probabilities for cell interior, borders, and background -- in
line with our multiclass approach -- and then minimized an energy cost function
to match the class probabilities via graph--cuts. We favored to avoid the
pitfalls of graph-cuts and the thresholding adopted in their formulation to
define seeds within cells.  Chen {\it et al.} \cite{chen2016} proposed DCAN, a
contour aware FCN to segment glands from histology images towards improving the
automatic diagnosis of adenocarcinomas. They also modeled a loss with contours
which led them to win the {\it 2015 MICCAI Gland Segmentation Challenge} \cite{gland2017}
confirming the advantages of explicitly learning contours.  Recently, Xu {\it
et al.} \cite{xu2017gland} proposed a three branch network to segment
individual glands in colon histology images.

Mask R-CNN \cite{he2017mask} is considered to be the state of the art in
instance segmentation for natural images. It classifies object bounding boxes
using Faster R-CNN \cite{ren2015} and then applies FCN inside each box to
segment a single object therein. Natural images, contrary to single channel,
low entropy cell microscope images, are much richer in information. Natural
objects in general differ to a great extent, making discrimination
comparatively easier.  Nevertheless, we plan to experiment and adapt Mask R-CNN
to our cell images after sufficient training data has been collected and
annotated.

{\bf Notation and definitions}.  We are given a training set
$S=\{(x_1,g_1),...,(x_N,g_N)\}$, with cardinality $|S| = N$, where
$x_k\colon\Omega\to\Re{}, \Omega\subset\Re{2}$, is a gray-level image and
$g_k\colon\Omega\to\{0,1\}$ its binary ground truth segmentation. Let $(x,g)$
be a generic tuple from $S$. We call $g^0$ and $g^1$, respectively, the
background and foreground subsets of $g$, and more generally $g^l = \{ p
\left|\right. c(p) = l, p\in\Omega\}$, where $c(p)$ returns the class assigned
to pixel $p$, $c\colon\Omega\to\{0,\ldots,C\}$. We write the pixel indicator
function $\indi{g^l}(p)$ simply as $y(p,l)$, i.e. $y(p,l) = 1$ if $p\in g^l$,
otherwise $y(p,l) = 0$. The connected components of $g$, $g^T = \{t_j | t_j\cap
t_i = \emptyset, j\neq i\}, \bigcup_j t_j = g^1$, are the non-empty masks for
all trainable cells in $x$.  For a mask $t$, $\Gamma_t$ represents its contour
and $h(t)$ gives its convex hull, also written as $h_t$. We say $\Gamma =
\bigcup_t\Gamma_t$ is the set of all contour pixels in $g$.  A mask admits a
skeleton $s(t)$, also written as $s_t$, which is its medial-axis. The distance
transform $\phi_g\colon\Omega\to\Re{}$ assigns to every pixel of $g$ the
Euclidean distance to the closest non-background pixel.  Touching cells in an
image $x$ share a common boundary (see \eg\ Fig.\ref{fig:cells}), which, by
construction, is a one pixel wide background gap separating their respective
connected components in $g^T$ (see figure in Algorithm \ref{alg:label3}).

\section{Multiclass and focus weights}
\label{sec:pmethod}
We propose higher weights to alleviate the imbalance of classes in the training
data and to emphasize cell contours, specially at touching borders, while
maintaining lower weights for the abundant, more homogeneous, easily separable
background pixels. However, it is also critical that background pixels around
cell contours should carry proportionally higher weights as they help capturing
cell borders more accurately specially in acute concave regions.

\begin{figure}[b!]
  \begin{center}
  \includegraphics[width=0.98\columnwidth]{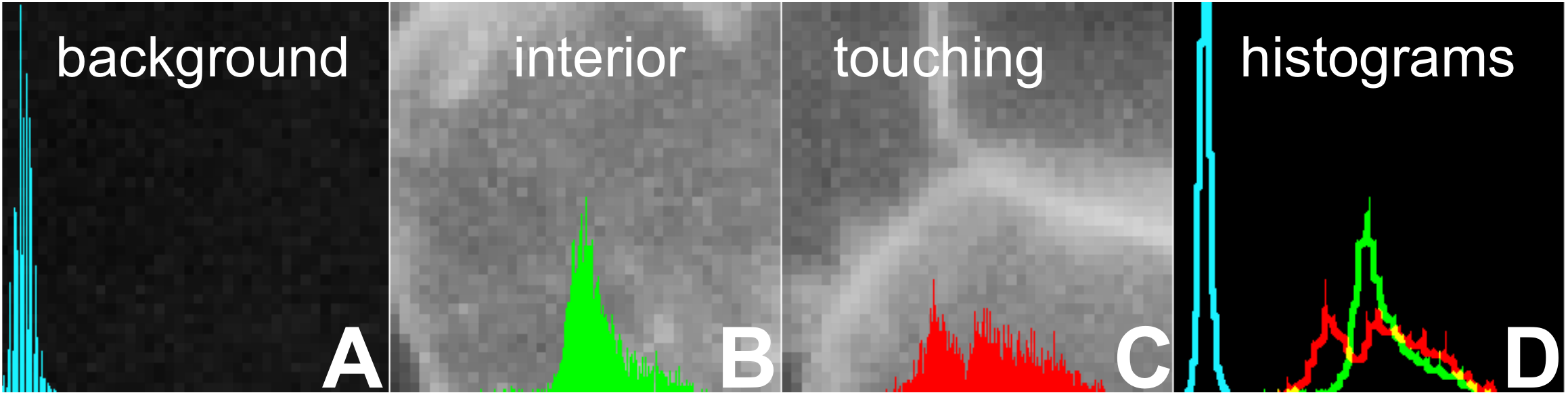}
  \caption{\footnotesize
  The distinct intensity and structural signatures of the three
    predominant regions -- background (A), cell interior (B), in-between cells
    (C) -- found in our images are shown above. Shown in panel (D)
    are the combined histogram curves for comparison. This distinction led us
    to adopt a multiclass learning approach which helped resolve the
    narrow bright boundaries separating touching cells, as seen in (C).}
  \vspace{-2.0mm}
    \noindent{
    \color{lightgray}\rule{1.0\columnwidth}{0.2mm}
    }
  \label{fig:regions}
  \end{center}\vspace{-0.8cm}
\end{figure}

Some authors, \eg\ \cite{ronneberger2015u,xu2017gland}, consider the one pixel
wide gaps in $g$ separating connected components to be part of the background
but with larger weights. By doing so, one might diminish the discriminative
power of the network as the foreground and background intensity distributions
overlap to some extent causing separation of pixels more difficult, as
suggested by the histograms shown in Fig.\ref{fig:regions}. There one can
notice the difference between the signatures of touching borders, cell
interiors, and background. If touching pixels are considered background pixels
for the purpose of training the network with only two classes, the distance
between the classes, foreground and background, would not be as pronounced as
if we have three separate classes.  This way, background is far off the other
two classes leaving interior and touching regions to be resolved, which is
helped with proper shape-aware weights. We believe, and show experimentally,
that by considering a multiclass learning approach we enhance the
discriminative resolution of the network and hence obtain a more accurate
segmentation of individual cells.

The goal of training our FCN network is to obtain a segmentation map $\hat{g}$
as close as possible to $g$, $\hat{g} \approx g$, given image $x$. When $x$ is
evaluated by a FCN a probability map $z\colon\Omega\to\Re{C}$ is obtained such
that $z(p)$ reports the probabilities of pixel $p$ belonging to each class. The
binary $\hat{g}$ can be obtained from $z$ applying a decision rule, like the
maximum {\it a posteriori} or, in our case, Algorithm\ref{alg:iassign}.

\vspace{-4mm}
\subsection{Class augmentation}
We perform label augmentation on the binary $g$ to create a third class
corresponding to touching borders. This is done using morphological operations
(Algorithm \ref{alg:label3}). By design, this new class occupies a slightly
thicker region than the original gap between cells. Training can now be done
using an augmented $g$ and the resulting map $z$ will have an extra class
representing the distribution for touching pixels.

{\noindent
\begin{minipage}{.7\columnwidth}
\begin{algorithm}[H]
\caption{Augment ground truth}\label{alg:label3}
\begin{algorithmic}[1]
\Procedure{labelAugment}{$g$,$s_e$}
\State ${g^{\prime}} \gets (g \oplus s_e) \ominus s_e$
\State ${g^{\prime}} \gets {g^{\prime}} - g$
\State ${g^{\prime}} \gets {g^{\prime}} \oplus s_e$
\State ${g} \gets {g} + (max(g) + 1)*{g^{\prime}}$
\EndProcedure
\end{algorithmic}
\end{algorithm}
\end{minipage}
\begin{minipage}{.28\columnwidth}
  \label{fig:alg1}
  \includegraphics[width=0.98\linewidth]{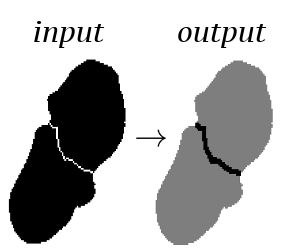}
   \footnotesize{$\oplus$ dilation $\ominus$ erosion}
\end{minipage}
}
\vspace{-2mm}
\fontdimen2\font=0.4ex
\begingroup
\captionof*{algorithm}{Mapping binary to three classes ground truth is done
using morphological operations. We use a $3\times3$ square structuring element
for $s_e$. Images are inverted for illustration purpose.}
\endgroup
\fontdimen2\font=\origiwspc %
\vspace{-2mm}
{\noindent\color{lightgray}\rule{1.0\columnwidth}{0.2mm}}
\subsection{Focus weights}
\begin{figure}[t!]
  \begin{center}
  \includegraphics[width=0.98\columnwidth]{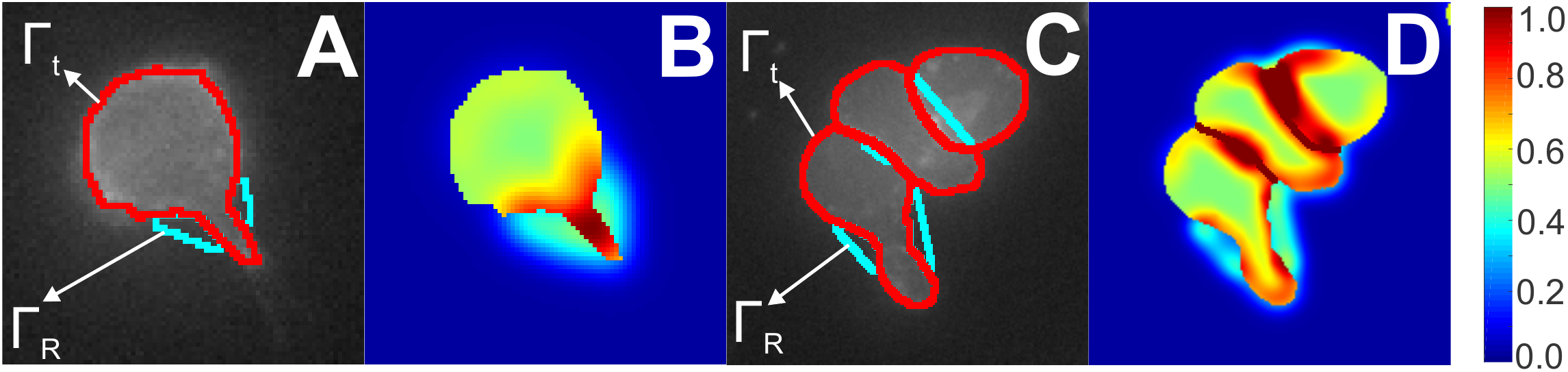}
  \caption{\footnotesize
  SAW for single (B) and touching (D) cells. Contours $\Gamma_t$ are shown in red and concavities $r$ in cyan in panels (A) and (C). Color code is normalized to maximum weight value. Note the large weights (in red) on narrow and concave regions, which help the network learn to segment accurate contours.} 
  \vspace{-2.0mm}
    \noindent{\color{lightgray}\rule{1.0\columnwidth}{0.2mm}}
  \label{fig:hcwm}
  \end{center}\vspace{-1.0cm}
\end{figure}

\begin{figure}[b!]
  \begin{center}
  \includegraphics[width=0.98\columnwidth]{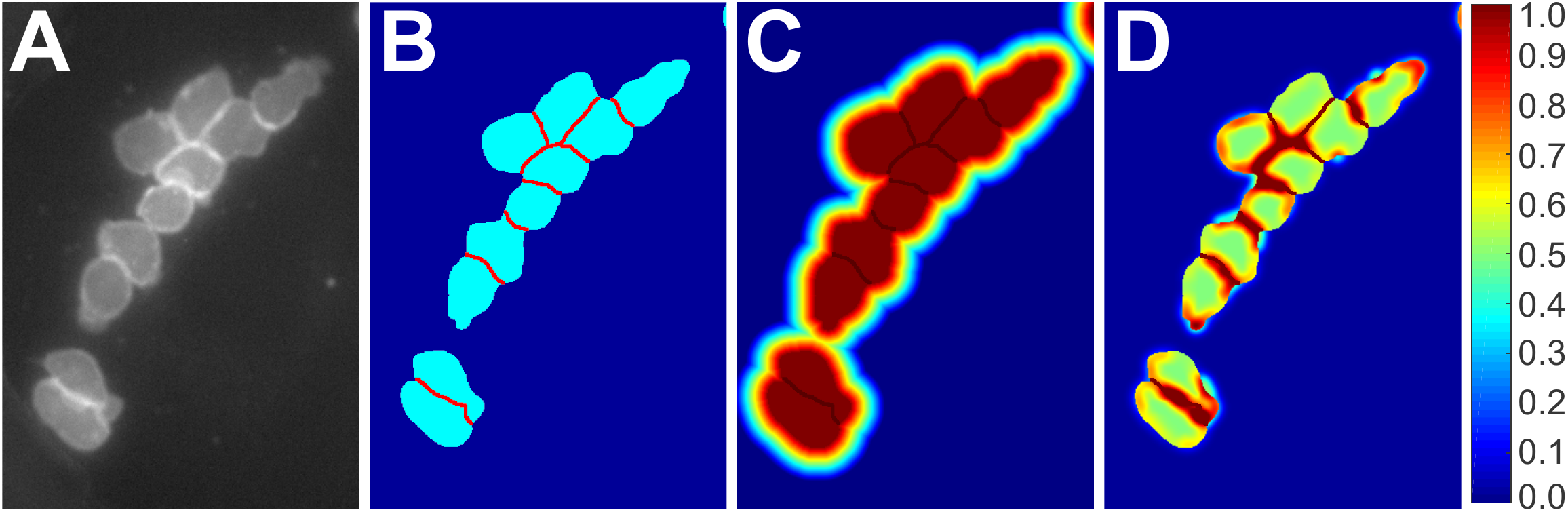}
  \caption{\footnotesize
  An example of cluttered cells is shown in panel (A). The weight maps, from 
  left to right are the plain class balancing weight map (B), our proposed
  distance transform based Weight map (C), and our shape aware weight map (D). Color code is normalized to maximum weight value with reds representing higher weights and blue small weights.}
  \vspace{-2.0mm}\noindent{\color{lightgray}\rule{1.0\columnwidth}{0.2mm}}
  \label{fig:wmaps}
  \end{center}\vspace{-0.8cm}
\end{figure}

The weighted cross entropy loss function \cite{ronneberger2015u} is used
to focus learning on important but underrepresented parts of an image:
\vspace{-2mm}
\begin{equation}
\vspace{-1mm}
L(y, z) = - \sum_{p \in \Omega}\sum_{l=0}^{C}  w(p,\theta)y(p,l) \log \text{smax}_l(z(p))
\label{eq:wceloss}
\end{equation}
where $\text{smax}_l(u)= {exp(u_{l})}/{\sum_{j=0}^{C} exp(u_j)}$ is the softmax
function applied to vector $u$, $\log$ is the logarithm function, $w(p,\theta)$
is a known weight at $p$ parameterized by $\theta$, $y(p,l)$ is the class
indicator function, $z(p)$ is the unknown feature vector for pixel $p$, and $C
= 2,3$.  We propose a distance transform based weight map (DWM)
\begin{equation}
w^{DWM}(p,\beta)= w_0(p)\left(1 - \min \left( \phi_g(p)/\beta , 1 \right)\right) 
\label{eq:dwm}
\end{equation}
where $\beta \geqslant 1$ is a control parameter that decays the weight from
the contour, and $w_0(p) = 1/|g^{c(p)}|$ is the class imbalance weight
\cite{ronneberger2015u}, inversely proportional to the number of pixels in the
class. Typically $|g^0| > |g^1| > |g^2|$, but the weights hold
regardless. Note that $w^{DWM}$ vanishes for $\phi > \beta$, and 
for $\phi \in [0,\beta]$, we have a linear decay $w^{WDM}(p) = w_0(p)(1 -
\phi_g(p)/\beta)$ for background pixels $p\in g^0$. Non-background pixels ($\phi = 0$) have
class constant $w_0$ weights. Fig.\ref{fig:wmaps}C shows $w^{DWM}$ for $\beta=30$.
It turns out that segmenting valid minutiae (\eg\ cell tip in
Fig.\ref{fig:hcwm}A,B), usually in the form of high curvature and narrow
regions, requires stronger weights. This led us to formulate a shape aware
weight map to take into account small but important nuances around contours.

The concave complement of $t\in g^T$ is $r(t) = h(t)\setminus{t}$.  Let $K$
be a binary image with skeletons $s(t)\cup s(r)$ as foreground pixels, and $\phi_{K}$ the
distance transform over $K$. 
We call $\Gamma_{H}=\Gamma_t \cup \Gamma_r$. Our shape aware weight map (SAW) is
\begin{equation}
w^{SAW}(p,\tau,\sigma) = w_0(p) + F_{\sigma} \ast w_c(p,\tau)
\label{eq:hcwm}
\end{equation}
where convolution with filter $F_{\sigma}$, which combines copy padding and
Gaussian smoothing, propagates $w_c$ values from $\Gamma_H$,
\begin{equation}
w_c(p,\tau)=\begin{cases}
1 -\phi_{K}(p)/\tau & \text{for } p \in \Gamma_{H} \\0 & \text{otherwise}
\end{cases}
\label{eq:hcwm1}
\end{equation}
to neighboring pixels. $\tau=\sup_{p\in\Gamma_H }\phi_{K}(p)$ is a distance normalization factor.
$w_c$ measures complexity for each $t$ by computing distances to the skeletons
of the mask and of its complement to assess how narrow are the regions around
the contours. The smallest distances give rise to larger weights. The value of
$\tau$ governs the distance tolerance and it is application dependent. Note that
SAW assigns large weights to small objects without any further processing or
loss function change contrary to what has been proposed by Zhou et al.
\cite{zhou2017focal}. Examples of SAW for single and touching cells are shown
in Fig.\ref{fig:hcwm} and comparatively for a cluster in Fig.\ref{fig:wmaps}D.
\subsection{Assigning touching pixels}
\label{sec:assign}
The touching pixels in the network generated probability map $z$ need to be distributed to adjacent cells. We do this in Algorithm\ref{alg:iassign} by assigning each pixel $p$, for which it has been determined that $p\in\hat{g}^2$, to its closest adjacent cell. The method uses map $z$ and two given thresholds $\gamma_1$ and $\gamma_2$ as decision rules to build the final binary segmentation $\hat{g}$: $\hat{g}^1$ contains the segmented cell masks $\hat{g}^T$ in background $\hat{g}^0$. The threshold $\gamma_1$ is used to determine touching pixels and $\gamma_2$ to determine cell masks: $z_2(p) > \gamma_1\rightarrow p\in\hat{g}^2$, and 
$z_1(p) > \gamma_2\rightarrow p\in\hat{g}^1$. All other pixels are background.
\begin{algorithm}[t!]
\caption{Pixel class assignment}\label{alg:iassign}
\begin{algorithmic}[1]
\Procedure{instanceAssign}{$z$,$\gamma_1$,$\gamma_2$}
\State $\textbf{if } z_2(p)>\gamma_1 \textbf{ then } p \in \hat{g}^2$
\State $\textbf{if } z_1(p)>\gamma_2 \textbf{ and } p \notin \hat{g}^2 \textbf{ then } p \in \hat{g}^1$
\For{$\textbf{all } p \textbf{ such that } p \in \hat{g}^2$}
\State $q \gets \arg \min_{q \in \hat{g}^1} ||p  - q||^{2}_{2}$
\State $\hat{g}(p) \gets \hat{g}(q)$
\State $t \gets t \cup \{p\}, q \in t, t \in \hat{g}^T$
\EndFor
\EndProcedure
\end{algorithmic}
\end{algorithm}
\vspace{-2mm}
\section{Results}
\label{sec:resul}
\vspace{-2mm}
We demonstrate our method in a manually curated \tcell\ segmentation
dataset containing thirteen images of size 1024x1024. We augmented this data
with warping and geometrical transformations (rotations, random crops,
mirroring, and padding) in every training iteration. Ten images were used for
the U-Net training \cite{ronneberger2015u}. We call UNET2 the use of
U-Net with two classes and weights from \cite{ronneberger2015u}. The same model with label
augmentation is referred as UNET3. DWM and SAW refer to training with U-Net network
using the proposed $w^{DWM}$ and $w^{SAW}$ weights, respectively. We refer to
FL as the focal loss work in \cite{zhou2017focal} which was applied in the
segmentation of small objects using an adaptive weight map. We use its loss
combined with label augmented ground-truth. All networks were equally
initialized with the same normally distributed weights using the Xavier method
\cite{glorot2010understanding}. After training, binary segmentations are
created using the pixel assignment algorithm described in section
\ref{sec:assign}.
\begin{table}[!h]
\resizebox{\linewidth}{!}{%
\begin{tabular}{lcccccc}
\footnotesize
Radius & 2 & 3& 4 &5 &6 &7\\\hline
\multicolumn{7}{c}{Training set}\\
UNET2 & 0.7995        & 0.8762 & 0.8936 & 0.9053 & 0.9109 & 0.9137\\
UNET3 & 0.7997        & 0.8896 & 0.9087 & 0.9244 & 0.9320 & 0.9356\\
FL    &0.7559         &0.8557 & 0.8821 & 0.9007 & 0.9087 & 0.9125\\
DWM   &0.8285         &0.9139 & 0.9333 & 0.9484 & \textbf{0.9546} & \textbf{0.9578}\\
SAW  &\textbf{0.8392}&\textbf{0.9183}& \textbf{0.9353} & \textbf{0.9485} & 0.9544 & 0.9573\\
\hline
\multicolumn{7}{c}{Testing set}\\
UNET2 & 0.6158 & 0.7116 & 0.7368 & 0.7627 & 0.7721 & 0.7828\\
UNET3 & 0.6529 & 0.7505 & 0.7770 & 0.8021 & 0.8158 & 0.8238\\
FL &0.5434&0.6566 & 0.6958 & 0.7263 & 0.7414 & 0.7505\\
DWM &0.6749&0.7847& 0.8156 & 0.8398 & 0.8531 & 0.8604\\
SAW&\textbf{0.7332}&\textbf{0.8298}& \textbf{0.8499} & \textbf{0.8699} & \textbf{0.8800} & \textbf{0.8860}\\
\end{tabular}
}
\vspace{-1.0mm}
\noindent{\color{lightgray}\rule{1.0\columnwidth}{0.2mm}}
\caption{\footnotesize
Results of the F1 score for different contour uncertainty radii. Our SAW method performed better than others, with DWM the second best on training data.}
\label{tab:segres}
\end{table}%

\begin{figure}[b!]
 \begin{center}
 \centering
\setlength{\tabcolsep}{1pt}
\begin{tabular}{cc}
\includegraphics[width=.5\linewidth]{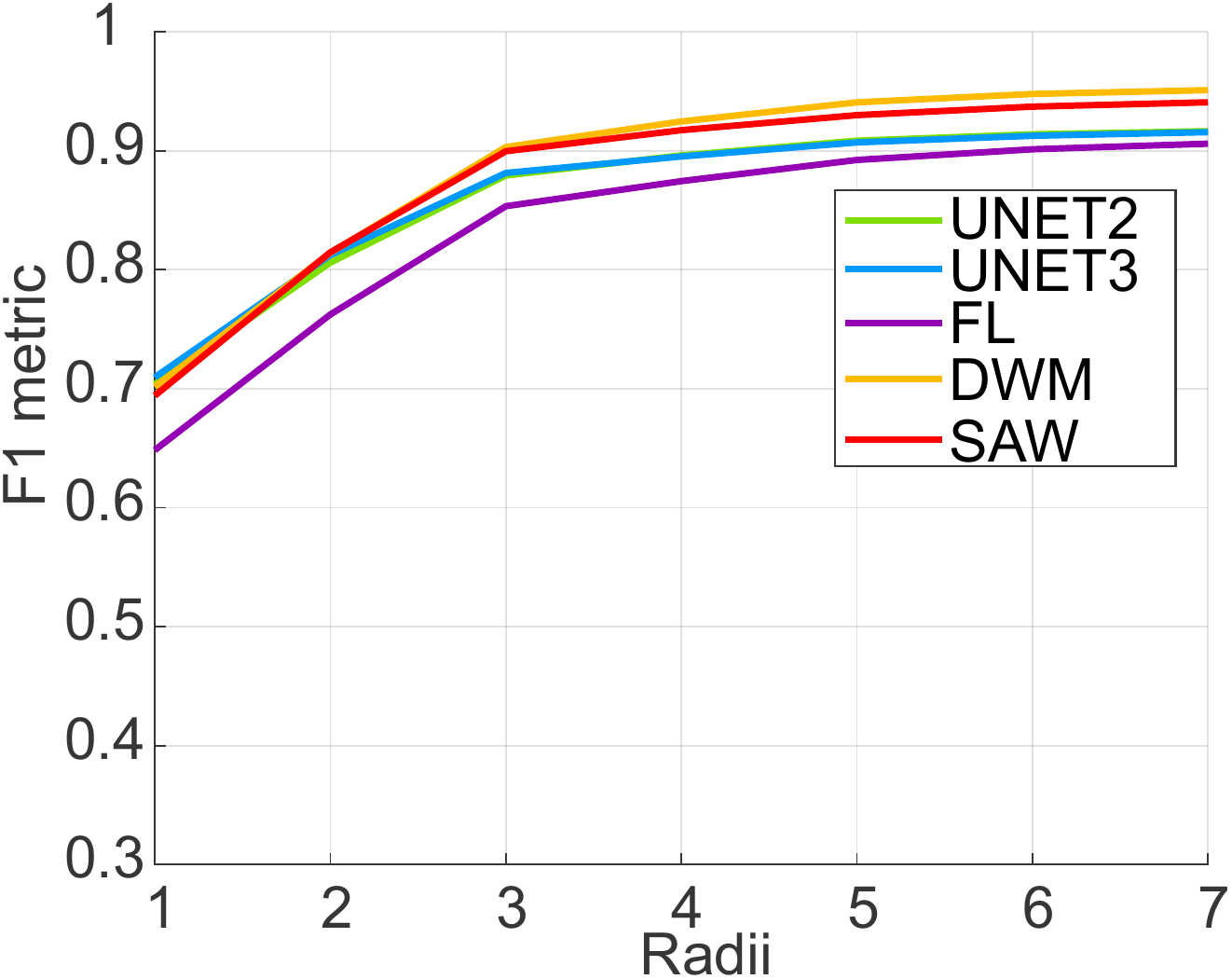}&
\includegraphics[width=.5\linewidth]{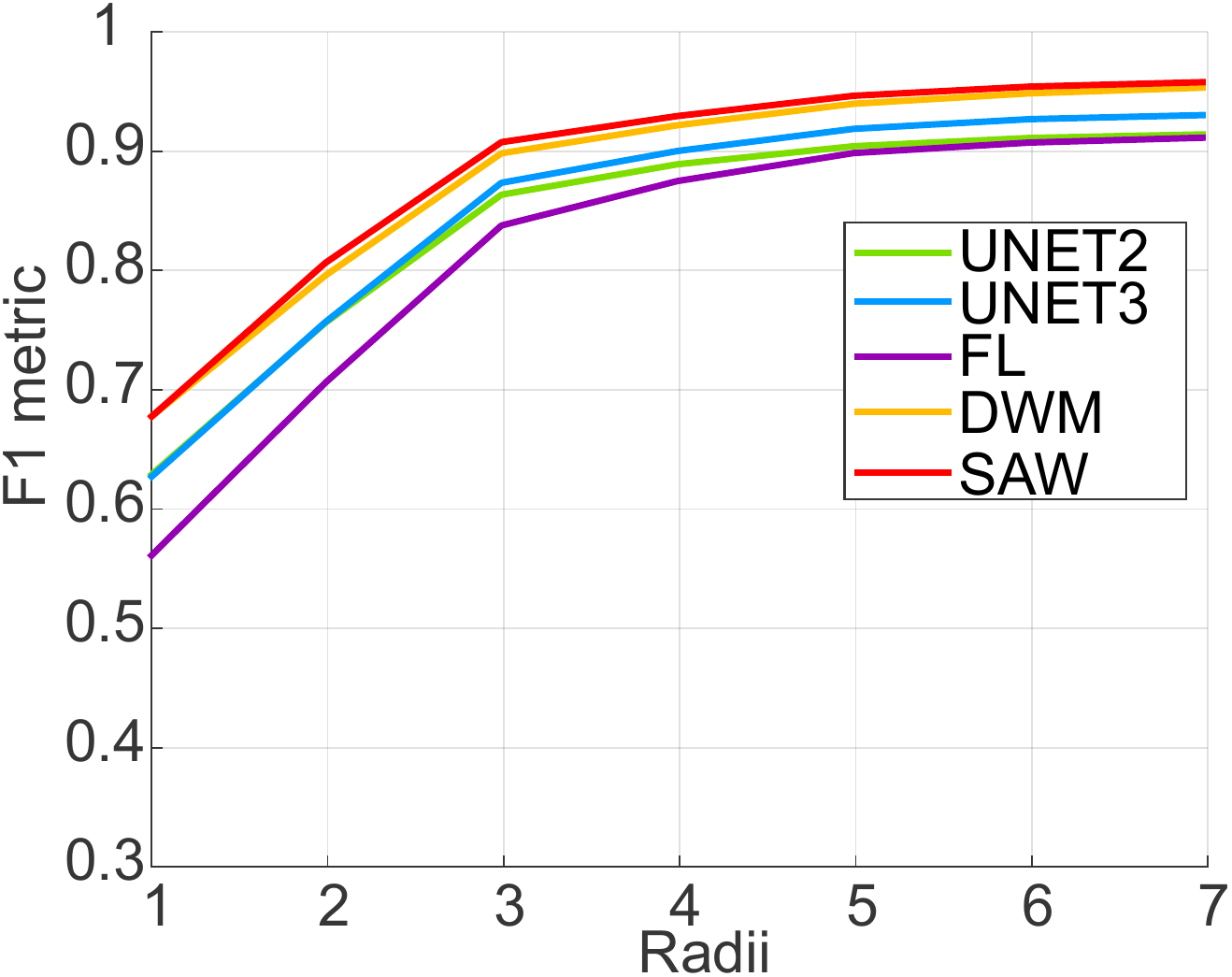}\\
1X field of view & 2X field of view\\ 
\includegraphics[width=.5\linewidth]{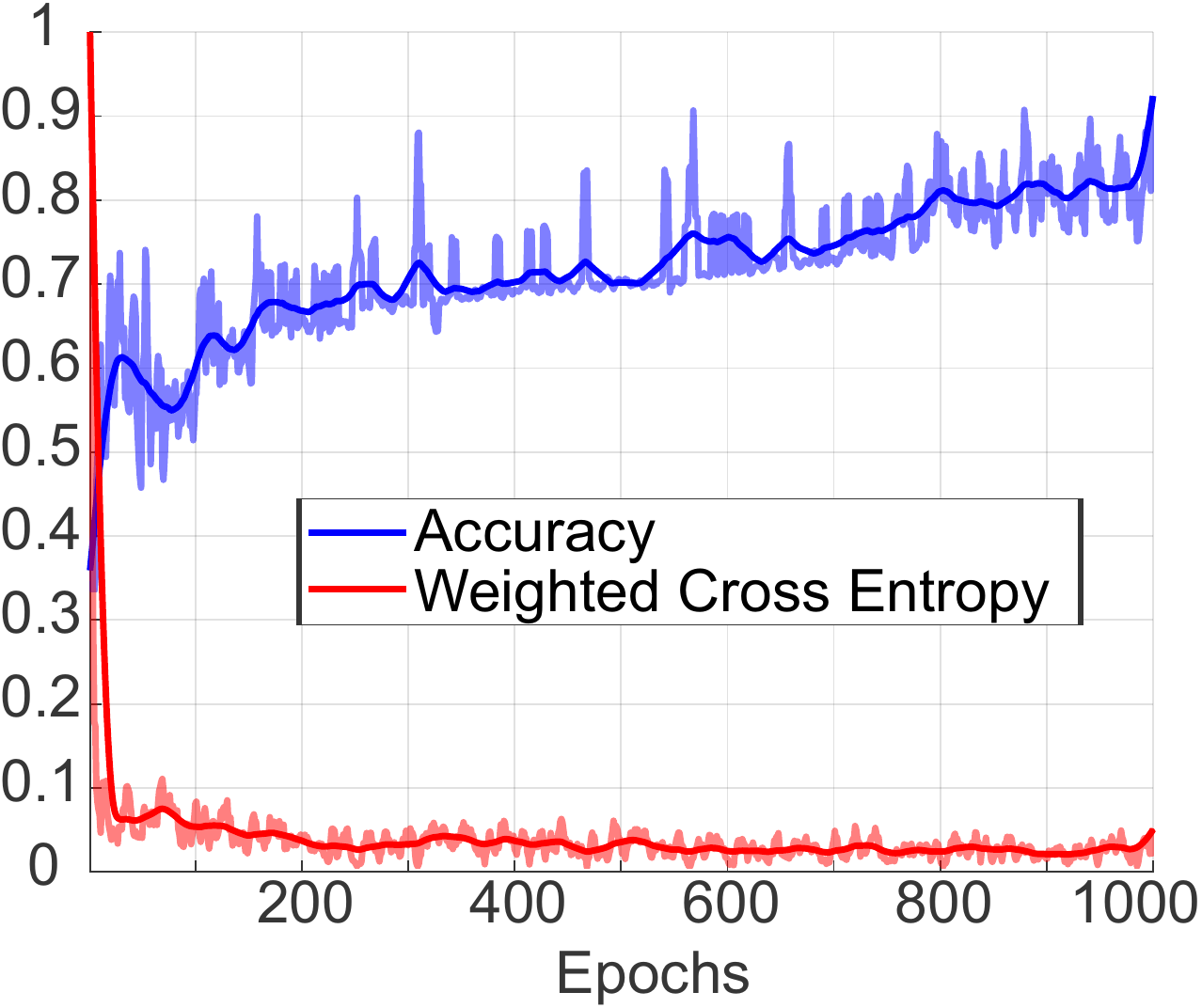}&
\includegraphics[width=.5\linewidth]{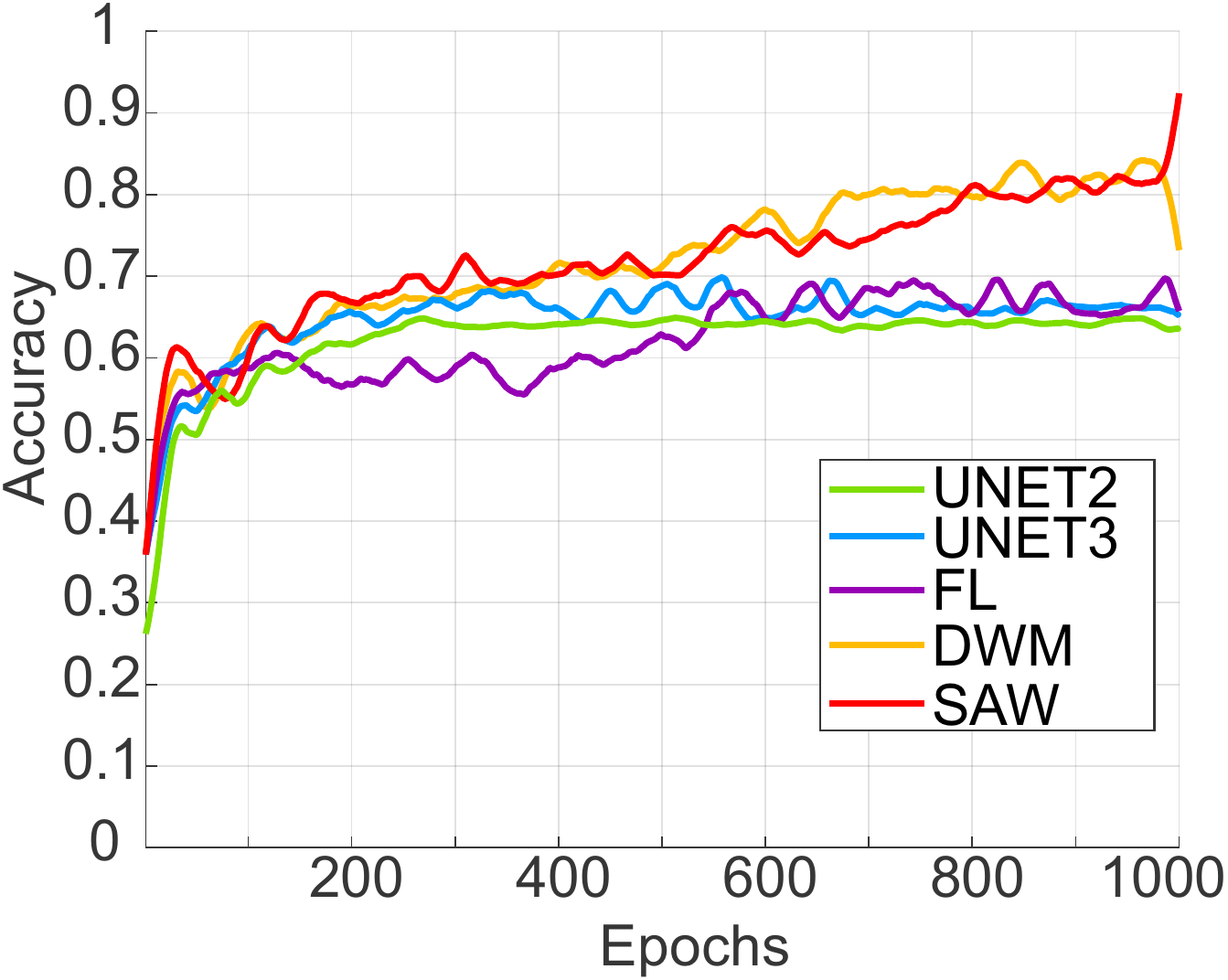}\\
\end{tabular}
 \caption{\footnotesize {\bf Top row}. F1 scores for radii $\rho \in [1,7]$ in 1X
 and 2X field of view size for each model. F1 values were consistently better
 for SAW and DWM.
 {\bf Bottom row}.  On the let panel, we show for all training epochs of the
 SAW network the class weighted accuracy (blue) and the weighted cross entropy
 (red).  In right panel we show the accuracy
 during training for all tested models, with outperforming rates for our DWM
 and SAW.}
 \vspace{-2.0mm}\noindent{\color{lightgray}\rule{0.95\columnwidth}{0.2mm}}
 \label{fig:radiivsf1} \end{center}\vspace{-0.8cm}
\end{figure}

We adopted the F1 score to compare computed contours to ground truth.  To allow
small differences in the location of contours, an uncertainty radius $\rho\in
[2,7]$, measured in pixels, is used for the F1 calculation, following
\cite{estrada2009benchmarking}. Table \ref{tab:segres} compares the results
from different methods for several {\it radii}. For all {\it radii} our
proposed methods outperform the other approaches. Better contour adequacy is
obtained mainly with SAW for $\rho < 6$ in the training set. In the testing phase,
however, higher generalization can be observed with SAW for all {\it radii}.
DWM was ranked second best. We will perform further tests with UNET3 to
increase separability but the accuracy we have achieved so far, see
Fig.\ref{fig:radiivsf1}, suggests improvements will not surpass DWM or SAW.

Plots of F1 score for different {\it radii} and fields of view are shown in
Fig.\ref{fig:radiivsf1} for all methods. We have experiemted with image sizes
1024x1024, 900x900, 500x500, and 250x250 corresponding to 1X, 1.1X, 2X and 4X
fields of view. Objects look smaller to the network when the image size is
reduced compromising instance segmentation. FL performed poorly when the
field of view is increased. In all cases the best performances were obtained
using SAW and DWM.  
To measure instance detection, every recognized cell with Jaccard Index
\cite{csurka2013good} greater than $0.5$ is counted as True Positive.
Contrary to the Intersection over Union (IoU) metric for detection
\cite{hosang2016makes} which uses bounding boxes, the Jaccard Index calculates
the instance adequacy from object segmentation. Precision, Recall and F1 are
calculated as described by Ozdemir {\it et
al.}\cite{ozdemir2010performance}. Table \ref{tab:detres} shows the instance
recognition metrics for all the approaches. The proposed methods outperform
with high margin all the other methods when the number of recognized instances
is taken into account. The SAW method showed an improvement of $6\%$ over
DWM for the training set and an improvement of $14\%$ for the testing set. 
UNET2 behaved poorly in cluttered cells, unable to separate then. We speculate
the combination of background and touching regions by UNET2 into a single class
prevented the proper classification of pixels.

These encouraging results suggest that combining multiclass
learning with pixel--wise shape aware weights might be advantageous to achieve
improved segmentation results. We will perform further tests with UNET3 to increase separability but the accuracy we have achieved so far suggests minor improvements.
\begin{table}[!h]
\resizebox{\linewidth}{!}{%
\begin{tabular}{lccccc}
\footnotesize
&UNET2 & UNET3 & FL & DWM & SAW\\\hline
\multicolumn{6}{c}{Training set}\\
Precision & 0.6506&0.7553 &0.7276&\textbf{0.8514}&0.8218\\
Recall    & 0.4187&0.6457 &0.4076&0.7191&\textbf{0.8567}\\
F1 metric & 0.5096&0.6962 &0.5225&0.7797&\textbf{0.8389}\\\hline
\multicolumn{6}{c}{Testing test}\\
Precision & 0.5546&0.7013 &0.6076&0.7046&\textbf{0.8113}\\
Recall    & 0.2311&0.3717 &0.2071&0.5195&\textbf{0.6713}\\
F1 metric & 0.3262&0.4858 &0.3089&0.5980&\textbf{0.7347}\\
\end{tabular}
}
\vspace{-1.0mm}
\noindent{\color{lightgray}\rule{1.0\columnwidth}{0.2mm}}
\caption{\footnotesize
Instance detection for Jacard Index above 0.5 is much pronunciated for
SAW meaning it can detect more cell instances than the other methods.}
\label{tab:detres}
\end{table}
\vspace{-6mm}
\section{Conclusions}
\label{sec:conclusion}
\vspace{-2mm}
We proposed two new shape based weight maps which improved the effectiveness of
the weighted cross entropy loss function in segmenting cluttered cells.  We
showed how learning with augmented labels for touching cels can benefit
instance segmentation. Experiments demonstrated the superiority of the proposed
approach when compared to other similar methods. In future work we will explore
learning procedures that adapt weights in the critical contour regions and
possibily improve results by training with more data.

\nocite{pena2018}

\bibliographystyle{IEEEbib}
\bibliography{refs}

\end{document}